\documentclass{article}
\usepackage{ijcai20}
\usepackage{todonotes}
\usepackage{enumitem}

\usepackage{times}

\usepackage{soul}
\usepackage{xcolor}
\usepackage[hyphens]{url}
\usepackage[hidelinks]{hyperref}
\usepackage[utf8]{inputenc}
\usepackage[small]{caption}
\usepackage{graphicx}
\usepackage{amsmath}
\usepackage{booktabs}
\usepackage{todonotes}

\newcommand{\rashkincorpus}{\texttt{TSHP-17}}
\newcommand{\proppycorpus}{\texttt{QProp} }
\newcommand{\emnlpcorpus}{\texttt{PTC} }
\newcommand{\mypar}{\subsection}

\title{A Survey on Computational Propaganda Detection}

\author{
Giovanni {Da San Martino}$^1$\footnote{Contact Author}\and
Stefano Cresci$^2$\and
Alberto Barr\'on-Cede\~no$^{3}$\and \\
Seunghak Yu$^{4}$\and
Roberto Di Pietro$^{5}$\And
Preslav Nakov$^{1}$
\\
\affiliations
$^1$Qatar Computing Research Institute, HBKU, Doha, Qatar\\
$^2$Institute of Informatics and Telematics, IIT-CNR, Pisa, Italy\\ 
$^3$DIT, Alma Mater Studiorum--Universit\`a di Bologna, Forl\`i, Italy\\
$^4$MIT Computer Science and Artificial Intelligence Laboratory,
Cambridge, MA, USA\\
$^5$College of Science and Engineering, HBKU, Doha, Qatar\\
\emails
\{gmartino, rdipietro, pnakov\}@hbku.edu.qa,
s.cresci@iit.cnr.it,
a.barron@unibo.it,
seunghak@csail.mit.edu
}

\begin{document}

\maketitle

\begin{abstract}

    Propaganda campaigns aim at influencing people's mindset with the purpose of advancing a specific agenda. They exploit the anonymity of the Internet, the micro-profiling ability of social networks, and the ease of automatically creating and managing coordinated networks of accounts, to reach millions of social network users with persuasive messages, specifically targeted to topics each individual user is sensitive to, and ultimately influencing the outcome on a targeted issue. 
    In this survey, we review the state of the art on computational propaganda detection from the perspective of Natural Language Processing and Network Analysis, arguing about the need for combined efforts between these communities. We further discuss current challenges and future research directions.
\end{abstract}

\section{Introduction\label{sec:introduction}}

The Web makes it possible for anybody to create a website or a blog and to become a news medium. Undoubtedly, this is a hugely positive development as it elevates freedom of expression to a whole new level, giving anybody the opportunity to make their voice heard. With the rise of social media, everyone can  reach out to a  very large audience, something that until recently was only possible for major news outlets.

However, this new avenue for self-expression has brought also unintended consequences, the most evident one being that the society has been left unprotected against potential manipulation from a multitude of sources. 
The issue became of general concern in 2016, a year marked by micro-targeted online disinformation and misinformation at an unprecedented scale, primarily in connection to Brexit and the US Presidential campaign; then, in 2020, the COVID-19 pandemic also gave rise to the first global infodemic.
Spreading disinformation disguised as news created the illusion that the information was reliable, and thus people tended to lower their natural barrier of critical thinking compared to when information came from different types of sources.

Whereas false statements are not really a new phenomenon ---e.g.,~yellow press has been around for decades--- this time things were notably different in terms of scale and effectiveness thanks to social media, which provided both a medium to reach millions of users and an easy way to micro-target specific narrow groups of voters based on precise geographic, demographic, psychological, and/or political profiling.

An important aspect of the problem that is often largely ignored is the mechanism through which disinformation is being conveyed, which is using \emph{propaganda techniques}.
These include specific rhetorical and psychological techniques, ranging from leveraging on emotions ---such as using \textit{loaded language}, 
\emph{flag waving},
\emph{appeal to authority},
\textit{slogans},
and \textit{clich\'{e}s}---
to using logical fallacies ---such as \textit{straw men}
(misrepresenting someone's opinion), \textit{red herring}
(presenting irrelevant data), \emph{black-and-white fallacy} 
(presenting two alternatives as the only possibilities), and \emph{whataboutism}.
Moreover, the problem is exacerbated by the fact that propaganda does not necessarily have to lie; it could appeal to emotions or cherry-pick the facts. Thus, we believe that specific research on propaganda detection is a relevant contribution in the fight against  online disinformation.

Here, we focus on {\em computational propaganda}, which is defined as ``propaganda created or disseminated using computational (technical) means'' \cite{Bolsover2017}. 
Traditionally, propaganda campaigns had been a monopoly of state actors, but nowadays they are within reach for various groups and even for individuals. 
One key element of such campaigns is that they often rely on coordinated efforts to spread messages at scale. Such coordination is achieved by leveraging botnets (groups of fully automated accounts)~\cite{zhang2016rise}, cyborgs (partially automated)~\cite{chu2012detecting} and troll armies (human-driven)~\cite{IRA:2018}, known as \emph{sockpuppets} \cite{Kumar:2017:AMS:3038912.3052677}, \emph{Internet water army} \cite{Chen:2013:BIW:2492517.2492637}, \emph{astroturfers} \cite{Ratkiewicz:2011:TMS:1963192.1963301}, and \emph{seminar users} \cite{SeminarUsers2017}.
Thus, a promising direction to thwart propaganda campaigns is to discover such coordination; this is demonstrated by recent interest by Facebook\footnote{\scriptsize\url{newsroom.fb.com/news/2018/12/inside-feed-coordinated-inauthentic-behavior/}} and Twitter\footnote{\scriptsize\url{https://help.twitter.com/en/rules-and-policies/platform-manipulation}}.

In order for propaganda campaigns to work, it is critical that they go unnoticed. This further motivates work on detecting and exposing propaganda campaigns, which should make them increasingly inefficient.
Given the above, in the present survey, we focus on computational propaganda from two perspectives: (\emph{i})~the content of the propaganda messages and (\emph{ii})~their propagation in social networks.

Finally, it is worth noting that, even though there have been several recent surveys on fake news detection \cite{Shu:2017:FND:3137597.3137600,DBLP:conf/wsdm/ZhouZSL19}, fact-checking \cite{thorne-vlachos-2018-automated}, and truth discovery \cite{Li:2016:STD:2897350.2897352}, none of them focuses on computational propaganda. 
There has also been a special issue of the Big Data journal on Computational Propaganda and Political Big Data \cite{Bolsover2017}, but it did not include a survey. Here we aim to bridge this gap.

\section{Propaganda}
\label{sec:propaganda}

The term \emph{propaganda} was coined in the 17th century, and initially referred to the propagation of the Catholic faith in the New World~\cite[p. 2]{Jowett:12}. It soon took a pejorative connotation, as its meaning was extended to also mean opposition to Protestantism. In more recent times, back in 1938, the Institute for Propaganda Analysis~\cite{InstituteforPropagandaAnalysis1938}, defined propaganda as ``\emph{expression of opinion or action by individuals or groups deliberately designed to influence opinions or actions of other individuals or groups with reference to predetermined ends}''. 

Recently, Bolsover et. al~\shortcite{Bolsover2017} dug deeper into this definition identifying its two key elements:
(\emph{i})~trying to influence opinion, and (\emph{ii}) doing so on purpose. 
Influencing opinions is achieved through a series of rhetorical and psychological techniques. 
Clyde R. Miller in 1937 proposed one of the seminal categorizations of propaganda, consisting of seven devices~\cite{InstituteforPropagandaAnalysis1938}, which remain well accepted today~\cite[p.237]{Jowett:12}: \emph{name calling}, \emph{glittering generalities}, \emph{transfer}, \emph{testimonial}, \emph{plain folks}, \emph{card stacking}, and \emph{bandwagon}. 
Other scholars consider categorizations with as many as eighty-nine techniques~\cite{Conserva:03}, and Wikipedia lists about seventy techniques.\footnote{\scriptsize\url{http://en.wikipedia.org/wiki/Propaganda_techniques}} However, these larger sets of techniques are essentially subtypes of the general schema proposed in~\cite{InstituteforPropagandaAnalysis1938}.

Propaganda is different from
\emph{disinformation}\footnote{\scriptsize\url{http://eeas.europa.eu/topics/countering-disinformation_en}}, in particular with reference to the truth value of the managed information and its goal, which in disinformation are  (\emph{i})~\emph{false},  and (\emph{ii})~\emph{intending to harm}, respectively. 
The (often-neglected) intention to harm popped up in 2016, due to both the Brexit referendum and the US Presidential elections, when society and academia discovered that the news cycle got \emph{weaponized} by disinformation. 
Contrarily, propaganda can hook to claims that are either true or false, and its intended objectives can be either harmful or harmless (even good\footnote{\scriptsize Think of Greta Thunberg's highly propagandistic speech at the UN in 2019.}). 
In practice, propaganda and disinformation are used synergetically to achieve specific objectives, effectively turning   social media into a weapon. 
Another related concept is that of ``fake news'', where the focus is on a piece of information being factually false. 

Although lying and creating fake stories is considered as one of the propaganda techniques (some authors refer to it as ``black propaganda'' \cite{Jowett:12}), there are contexts where this course of actions is often done without pursuing the objective to  influence the audience, as in satire and clickbaiting. These special cases are of less interest when it comes to fighting the weaponization of social media, and are therefore considered out of the scope for this survey.

\section{Text Analysis Perspective\label{sec:text}}

Research on propaganda detection based on text analysis has a short history, mainly due to the lack of suitable annotated datasets for training supervised models. 
There have been some relevant initiatives, where expert journalists or volunteers analyzed entire news outlets, which could be used for training. For example, \emph{Media Bias/Fact Check} (MBFC)\footnote{\scriptsize\url{http://mediabiasfactcheck.com}} is an independent organization analyzing media in terms of their factual reporting, bias, and propagandist content, among other aspects. 
Similar initiatives are run by \emph{US News \& World Report}\footnote{\scriptsize\url{www.usnews.com/news/national-news/articles/2016-11-14/avoid-these-fake-news-sites-at-all-costs}} 
and the European Union.\footnote{\scriptsize\url{www.usnews.com/news/national-news/articles/2016-11-14/avoid-these-fake-news-sites-at-all-costs};~~\url{http://euvsdisinfo.eu/}} 
Such data has been used in distant supervision approaches~\cite{mintz-etal-2009-distant}, i.e.,~by assigning each article from a given news outlet the label propagandistic/non-propagandistic using the label for that news outlet. 
Unfortunately, such coarse approximation inevitably introduces noise to the learning process, as we discuss in Section~\ref{sec:lessons}.  

In the remainder of this section, we review current work on propaganda detection from a text analysis perspective. This includes the production of annotated datasets, characterizing entire documents, and detecting the use of propaganda techniques at the span level.

\mypar{Available Datasets} Given that existing models to detect propaganda in text are supervised, annotated corpora are necessary. Table~\ref{tab:corpora} shows an overview of the available corpora (to the best of our knowledge),  with annotation both at the document and at the fragment level. 

Rashkin et al.~\shortcite{rashkin-EtAl:2017:EMNLP2017} released \rashkincorpus, a balanced corpus with document-level annotation including four classes: \emph{trusted}, \emph{satire}, \emph{hoax}, and \emph{propaganda}. \rashkincorpus~ belongs to the collection of datasets annotated via distant supervision: an article is assigned  to one of the classes if the outlet that published it is labeled as such by the \textit{US News \& World Report}. The documents were collected from the English Gigaword and from seven unreliable news sources.

\begin{table}
 \footnotesize
 \centering
\begin{tabular}{llc@{\hspace{1mm}} c@{\hspace{1mm}}cr}
\toprule
 \bf Corpus     & \bf Level & \bf Sources   & \bf Classes   & \bf Articles	&   \bf Prop.  \\ 
 \midrule
 \rashkincorpus & document &\,\,\,11\,\,\, (2)&  \,\,\,4  & 22,580    & 5,330\\
 \proppycorpus  & document & 104 (10)         &  \,\,\,2  & 51,294    & 5,737\\
 \emnlpcorpus   & text span & \,\,\,49 (13)   & 18        & \,\,\,\,\,\,451 & 7,385\\
 \bottomrule
\end{tabular}
 \caption{Textual datasets available to train supervised propaganda identification models at different granularity levels.}
\label{tab:corpora}
\end{table}

According to Barr\'on-Cede\~no et al.~\shortcite{BARRONCEDENO20191849} the low amount of sources considered per class is a downside of \rashkincorpus, as the systems trained on it might be modeling the news outlets, rather than propaganda itself (or any of the other three classes). To cope with this limitation, Barr\'on-Cede\~no et al. released \proppycorpus a twice-as-big binary imbalanced dataset in which $\sim 10\%$ of the articles belong to class \emph{propaganda}. 

Once again, the annotation in \proppycorpus is obtained by distant supervision; this time with information from MBFC.
Aside the binary propaganda \textit{vs.} trustworthy annotation, in \proppycorpus each article has associated metadata from its source such as the bias level (e.g., \emph{left}, \emph{center}, \emph{right}) from MBFC and geographical information, average sentiment, publication date, identifier, author, and official source name from GDELT.\footnote{\scriptsize\url{https://www.gdeltproject.org/}}

\begin{figure}[tbh]
    \setlength{\fboxsep}{0.75pt}
    \fbox{\includegraphics[width=0.48\textwidth]{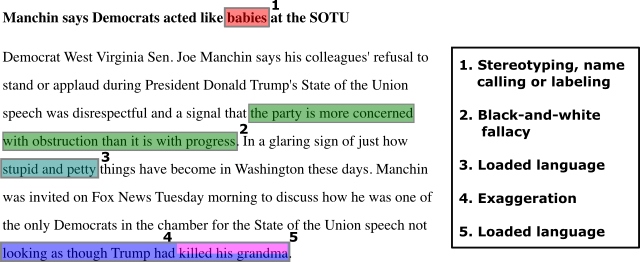}}
    \caption{Text excerpt with annotated propaganda techniques. \label{fig:example}}
\end{figure}

However, both \rashkincorpus~and \proppycorpus lack information about the precise location of a propagandist snippet within a document. 
Since propaganda is conveyed by using specific rhetoric and psychological techniques, a separate line of research recently aimed to identify the use of such techniques. In particular, Da San Martino et al.~\shortcite{EMNLP19DaSanMartino} proposed a dataset with assets that previously available resources lacked. First, their \emnlpcorpus corpus is manually judged by professional annotators, rather than using distant supervision. Second, the annotation is at the fragment level: specific text spans are flagged, rather than full documents. Third, it goes deeper into the types of propaganda, considering 18 propaganda techniques, rather than the binary \emph{propaganda} \textit{vs.} \emph{non-propaganda} setting. The curated list of techniques is summarized in Table~\ref{tab:prop_classes}. 
Whereas the volume of \emnlpcorpus is way lower than that of \rashkincorpus and \proppycorpus ---a few hundred articles against thousands--- it contains more than 7,000 propagandist snippets. See Figure~\ref{fig:example} for an example with annotations. 

Another relevant line of research is on computational argumentation, which deals with some logical fallacies considered to be propaganda techniques. 
 \citeauthor{Habernal.et.al.2017.EMNLP}~\shortcite{Habernal.et.al.2017.EMNLP} described a corpus with $1.3k$ arguments annotated with five fallacies such as \textit{ad hominem}, \textit{red herring}, and \textit{irrelevant authority}.

\mypar{Text Classification} 

Early approaches to propaganda identification are fairly aligned to the produced corpora. Rashkin et al.~\shortcite{rashkin-EtAl:2017:EMNLP2017} defined a classical four-classes text classification task: propaganda \textit{vs} trusted \textit{vs} hoax \textit{vs} satire, using the \rashkincorpus\, dataset.
Using word $n$-gram representation with logistic regression, they found that their model performed well only on articles from sources that the system was trained on.

\begin{table}[tbh]
\centering
\footnotesize
\begin{tabular}{@{}p{3cm}p{5cm}@{}}  
\toprule
\multicolumn{1}{l}{\bf Technique}  & \multicolumn{1}{c}{\bf Definition} \\
\midrule
Name calling & attack an object/subject of the propaganda with an insulting label \\\hline
Repetition & repeat the same message over and over\\\hline
Slogans & use a brief and memorable phrase \\\hline
Appeal to fear& support an idea by instilling fear against other alternatives \\\hline
Doubt & questioning the credibility of someone/something\\\hline
Exaggeration/minimizat. & exaggerate or minimize something \\\hline
Flag-Waving & appeal to patriotism or identity\\\hline
Loaded Language & appeal to emotions or stereotypes \\\hline
Reduction ad hitlerum & disapprove an idea suggesting it is popular with groups hated by the audience\\\hline
Bandwagon & appeal to the popularity of an idea\\\hline
Casual oversimplification & assume a simple cause for a complex event \\\hline
Obfuscation, intentional vagueness & use deliberately unclear and obscure expressions to confuse the audience\\\hline
Appeal to authority & use authority's support as evidence \\\hline
Black\&white fallacy & present only two options among many\\\hline
Thought terminating clich\'es & phrases that discourage critical thought and meaningful  discussions \\\hline
Red herring & introduce irrelevant material to distract \\\hline
Straw men & refute argument that was not presented \\\hline
Whataboutism & charging an opponent with hypocrisy\\
\bottomrule
\end{tabular}
\caption{List of the 18 propaganda techniques and their definitions. \label{tab:prop_classes}}
\end{table}

Barr\'on et al.~\shortcite{BARRONCEDENO20191849} used a binary classification setting: detecting propaganda \textit{vs} non-propaganda and experimented on \rashkincorpus and \proppycorpus corpora. 
They ran a massive set of experiments, investigating various representations, from writing style and readability level to the presence of certain keywords, together with logistic regression and SVMs, and confirmed that using distant supervision, in conjunction with rich representations, might encourage the model to predict the source, rather than to discriminate propaganda from non-propaganda. 

They advocated for providing assurance that  
test data come from news sources that were not used for  training, and investigated what representations remain robust in such a setting.

\mypar{Detecting the Use of Propaganda Techniques} 

Da San Martino et al.~\shortcite{EMNLP19DaSanMartino} defined two tasks, based on annotations from the \emnlpcorpus dataset: 
(\textit{i})~binary classification ---given a sentence in an article,  predict whether any of the 18 techniques has been used in it; (\textit{ii})~multi-label multi-class classification and span detection task ---given a raw text, identify both the specific text fragments where a propaganda technique is being used as well as the type of technique.
Such a fine-grained level of analysis may provide support and explanations to the user on why an article has been judged as propagandistic by an automatic system. 
The authors proposed a multi-granularity deep neural network that modulates the signal from the sentence-level task to improve the prediction of the fragment-level classifier.

A shared task was held within the \textit{2019 Workshop on NLP4IF: censorship, disinformation, and propaganda}\footnote{\scriptsize\url{http://www.netcopia.net/nlp4if/2019/}}, based on the \emnlpcorpus corpus and the task definitions above. 
The best-performing models for both tasks used BERT-based contextual representations.
Other approaches used contextual representations based on RoBERTa, Grover, and ELMo, or context-independent representations based on lexical, sentiment-based, readability, and TF-IDF features. Ensembles were also popular. 
Further details are available in the shared task overview paper~\cite{da-san-martino-etal-2019-findings}.

\section{Network Analysis Perspective}
\label{sec:relatedpropagandanetwork}

As seen in Section~\ref{sec:text}, the rhetoric techniques used for influencing readers' opinions can be detected directly in the text. Contrarily, identifying the intent behind a propaganda campaign requires analysis that goes beyond individual texts, involving (among others) classification of the social media users that contributed to injecting and spreading propaganda within a network. Thus, a necessary condition to detect the intention to harm implies detecting malicious coordination (i.e.,~coordinated inauthentic behavior). Throughout the years, this high-level task has been tackled in different ways.

\mypar{Early Approaches} Early approaches for detecting malicious coordination were based on classifying individual nodes in a network as either \emph{malicious} or \emph{legitimate}. Then, clusters of malicious nodes were considered to be acting in coordination. In other words, the concept of coordination was not embedded within the models, but it was added ``a posteriori''.
The vast majority of these approaches are based on supervised machine learning and each account under investigation was analyzed \textit{in isolation}. That is, given a group of accounts to analyze, the supervised technique was separately applied to each account of the group, that in turn received a label assigned by the detector.

The key assumption of this body of work is that each malicious account has features that make it clearly distinguishable from legitimate ones. This approach to the task also revolved around the application of off-the-shelf, general-purpose classification algorithms. Widely used algorithms include decision trees and random forests, SVMs, boosting and bagging (e.g.,~Adaptive Boost and Decorate) and, more recently, deep neural networks~\cite{kudugunta2018deep}.

The most widely known example of this kind of detectors, is Botometer~\cite{yang2019arming}, a social bot detection system. By leveraging more than 1,200 features for a social media account, it evaluates profile characteristics, social network structure, the  produced content (including sentiment expressions), and temporal features. Botometer simultaneously analyzes multiple dimensions of suspicious accounts for spotting bots. 
Instead, other systems solely rely on network characteristics~\cite{yang2015votetrust}, textual content~\cite{rangel2019overview}, 
or profile information~\cite{lee2014early}. These latter systems are typically easier to game, since they only analyze a single facet of the complex, evolving behavior of bad online actors.

\mypar{Evolving Threats} Despite having achieved promising initial results, these early approaches had several limitations. First, the performance of a supervised detector strongly depends on the availability of a ground truth (training) dataset. In most cases, a real ground truth is lacking and the labels are manually given by human operators. Unfortunately, as of 2020, we still have diverse and conflicting definitions of what a malicious account really is~\cite{grimme2017social}, and humans have been proven to suffer from annotation biases and to largely fail at spotting sophisticated bots and trolls~\cite{cresci2017paradigm}. To make matters worse, it has been demonstrated that malicious accounts ``evolve'' (i.e.,~they change their characteristics and behaviors) in an effort to evade detection by established techniques~\cite{cresci2017paradigm}. Nowadays, sophisticated malicious accounts are using the same technological weapons as their hunters ---such as powerful AI techniques--- for generating credible texts (e.g.,~with GPT-2), profile pictures (e.g.,~with StyleGAN)\footnote{\scriptsize\url{https://www.wired.com/story/facebook-removes-accounts-ai-generated-photos/}}, and videos (e.g.,~using deepfakes), thus dramatically increasing their capabilities of impersonating real people, and hence of escaping detection.

\mypar{Modern Approaches} The difficulties at detecting sophisticated bots and trolls with early approaches lead to a new research trend whose primary characteristic is to target groups of accounts as a whole, rather than focusing on individual accounts. In recently proposed detectors, coordination is considered a key feature to analyze, and it is modeled within the detectors themselves.
The rationale for this choice is that malicious accounts act in coordination (e.g.,~sbots are often organized in botnets, trolls form so-called troll armies) to amplify their effect~\cite{zhang2016rise}.
Moreover, by analyzing large groups of accounts, modern detectors also have more data to exploit for fueling powerful AI algorithms~\cite{sun2017revisiting}. The shift from individual to group analysis was accompanied by another shift from general-purpose machine learning algorithms, to ad-hoc algorithms specifically designed for detecting coordination. In other words, the focus shifted from feature engineering to learning effective feature representations and of designing brand-new and customized algorithms~\cite{cai2017detecting}. Many modern detectors are also unsupervised or semi-supervised, to overcome the generalization deficiencies of supervised detectors that are severely limited by the availability of exhaustive training datasets~\cite{de2018lobo}.

Examples of such systems implement network-based techniques,  aiming at detecting suspicious account connectivity patterns~\cite{liu2017holoscope,chetan2019corerank,pacheco2020unveiling}. 
Coordinated behavior appears as near-fully connected communities in graphs, dense blocks in adjacency matrices, or peculiar patterns in spectral subspaces~\cite{jiang2016inferring}. 
Other techniques adopted unsupervised approaches for spotting anomalous patterns in the temporal tweeting and retweeting behaviors of groups of accounts ---e.g.,~by computing metrics of distance out of the accounts activity time series and by subsequently account clustering~\cite{chavoshi2016debot,mazza2019rtbust}. 

The rationale behind such approaches is based on evidence suggesting that human-driven and legitimate behaviors are intrinsically more heterogeneous than automated and inauthentic ones~\cite{cresci2019emergent}. Consequently, a large cluster of accounts with highly similar behavior might serve as a red flag for coordinated inauthentic behavior. Distance (or similarity) between account activity time series was computed via dynamic time warping~\cite{chavoshi2016debot}, or as the Euclidean distance between the feature vectors computed by an LSTM autoencoder~\cite{mazza2019rtbust}.
More recently, other authors investigated the usefulness of Inverse Reinforcement Learning (IRL) for inferring the intent that drives the activity of coordinated groups of malicious accounts. 
Inferring intent and motivation from observed behavior has been extensively studied in the framework of IRL, with the main goal of finding the rewards behind an agent's observed behavior. 
The inferred rewards can then be used as features in supervised learning systems aimed at detecting malicious and coordinated agents.

The switch from early to modern detectors demonstrated that the approach (e.g.,~individual \textit{vs} group-based, supervised \textit{vs} unsupervised) to the task of propaganda and malicious accounts detection can have serious repercussions on detection performance. However, some scientific communities naturally tend to favor a specific approach. For example, the majority of techniques that perform network analysis (e.g.,~by considering the social or interactions graph of the accounts) are intrinsically group-based. More often than not, they are also unsupervised. Contrarily, all techniques based on textual analyses, such as those that solely rely on natural language processing, are supervised detectors that analyze individual accounts~\cite{rangel2019overview}. As a consequence, some combinations of the cited approaches ---above all, text-based detectors that perform unsupervised group analysis--- are almost unexplored. For the future, it would thus be advisable to put efforts along the highlighted research directions that have been mostly overlooked until now.

\section{Lessons Learned\label{sec:lessons}}

The main lesson from our analysis is that there is a disconnection between NLP and Network Analysis communities when it comes to fighting Computational Propaganda, and therefore combined approaches may lead to systems significantly outperforming the current state of the art. A detailed analysis is reported in the following.

\mypar{Text Analysis Lessons} From a text analysis perspective, we see that there is a lack of a suitable dataset for document-level propaganda detection. The attempts to use distant supervision as a substitute, by projecting labels from media to all the articles they have published is problematic in many aspects, even when done carefully. Indeed, distant supervision inevitably introduces noise in the learning process, as it is based on the wrong assumption that all articles from a given source would be either propaganda  or non-propaganda.  In reality, a propagandist source could periodically post objective non-propagandist information to boost its credibility. 

Similarly, sources that are generally recognized as objective might occasionally post information that promotes a particular agenda. 
One way to deal with this issue might be to devise advanced learning algorithms, such as Generative Adversarial Networks (GANs), which can be trained to avoid specific biases, i.e.,~modelling the article source. Another issue with distant supervision is that while it is acceptable for training, it cannot give a fair assessment of a system at testing time, something that previous work has ignored.

Another lesson is that it seems more promising to focus on detecting the use of fine-grained propaganda techniques in text. Propaganda techniques are well-defined and well-known in the literature, and thus it makes sense to focus on them, as they are the very device on which propaganda is built. Notably, a proper dataset is already available for this new task, it is of reasonable size (350K tokens, which compares well to datasets for the related task of named entity recognition, whose typical size is 200K tokens), and covers a wide range of 18 commonly accepted techniques, comprising both various kinds of appeal to emotions as well as logical fallacies.

\mypar{Network Analysis Lessons} Typically, when scholars and OSN administrators identify new coordinated behavior that goes undetected by existing techniques, as a reaction they start the development of new detectors. The implication of this reactive approach is that improvements occur only some time after having collected evidence of a new mischievous behavior. Bad actors thus benefit from a large time span ---the time needed to design, develop, and deploy a new detector--- during which they are free to tamper with our online environments. 

A second lesson learned is related to the use of machine learning algorithms, the vast majority of which are designed to operate in environments that are stationary and neutral. Unfortunately, in the task of propaganda campaign detection both assumptions are easily violated, yielding unreliable predictions and severely decreased performance~\cite{goodfellow2014generative}. Stationarity is violated by the mechanism of evolution of malicious accounts, resulting in accounts exhibiting different behavior and characteristics over time. Neutrality is violated as well, since propaganda spreaders and bot masters are actively trying to fool detectors. Consequently, the exceptional results in malicious accounts detection that we reported in our papers might be actually largely exaggerated.

Adversarial machine learning may however mitigate both previous issues, since the existence of adversaries is accounted for by design. We could thus apply adversarial machine learning to study vulnerabilities of existing detectors and the possible attacks the cited vulnerabilities could lead to, before they have been exploited by the adversaries.
Interestingly, this paradigm has recently been applied for improving bot detection as well as for fake news detection~\cite{wu2020using,zellers2019defending}. Finally, it is worth noting that all tasks related to the detection of online deception, manipulation, and automation ---including, but not limited to, propaganda campaign detection--- are intrinsically adversarial.

\section{Challenges and Future Forecasting}
\label{sec:challenges}

\mypar{Major Challenges} Computational propaganda detection is still in its early stages and the following challenges need to be addressed:
\begin{enumerate}[noitemsep, leftmargin=*]
    \item Text is not the only way to convey propaganda. Sometimes, pictures convey stronger messages than texts, as for certain political memes. Thus, it is becoming increasingly necessary to analyze multiple modalities of data (e.g.,~images, videos, speech). This is challenging because, even if some research was conducted on how to effectively understand cross-modal information in various domains, little has been done on what information (provided by a given  modality) can be leveraged  to detect propaganda.
    
    \item Explainability is a desirable feature of propaganda detection systems in order to make them accepted at large. In fact, it is crucial to be able to motivate decisions, especially controversial ones (e.g.,~banning of OSN accounts or removal of posts/news). However, most of the recent developments in propaganda and coordination detection are based on deep learning, which lacks explanability ---for the short and medium term, at least.
    
    \item In addition to being able to classify individual documents as propaganda or single accounts as deceptive/coordinated, it would be useful to also provide information towards understanding the goals and the strategy of propaganda campaigns~\cite{atanasov-etal-2019-predicting}. This problem currently stands as largely unsolved and calls for joint efforts in propaganda and coordination detection.
    
    \item Recent advances in neural language models have made it difficult even for humans to detect synthetic text. 
    Zellers et al.~\shortcite{zellers2019defending} showed that a template system helps manipulate the output format of a language model, while Yang et al.~\shortcite{yang2018unsupervised} suggested how to transfer the style of the language model to the target domain. With all building blocks already in place, it is likely that automatically-generated propaganda will surface in the near future.

    \item The vast majority of existing detectors are evaluated only on a single annotated dataset. Often, the dataset is collected and annotated for a specific study, and is subsequently disregarded. As such, we currently lack the ability to evaluate detectors' capability  of generalizing the performance obtained \textit{in silico}, also when applied in-the-wild. 
     For the future, it is advisable to devote additional efforts to curate large annotated datasets. Extensive data sharing initiatives ---such as that of Twitter related to recent information operations\footnote{\scriptsize\url{http://transparency.twitter.com/en/information-operations.html}}--- are thus particularly welcome.
    
    \item When dealing with user-generated data, ethical considerations are also important. We should thus guarantee that all analysis and potential sharing of datasets are conducted respecting the privacy of the involved users. This can also affect data availability, as demonstrated by the Facebook/Social Science One URL dataset\footnote{\scriptsize\url{http://socialscience.one/blog/unprecedented-facebook-urls-dataset-now-available-research-through-social-science-one}}, whose release was postponed for almost two years due to the need to implement robust privacy-preserving mechanisms. 
\end{enumerate}

\mypar{Forecasting} Given the above challenges and the existence of some previously remarked  under-explored directions, we highlight the following research directions:

\begin{enumerate}[noitemsep, leftmargin=*]
    \item There is growing motivation for jointly tackling the textual and the network aspects of propaganda detection, as relying on a single paradigm is a recipe for failure. For instance, if a pre-trained language model such as GPT-2 is used as an automated propaganda generation method, it may become ineffective to detect propaganda when focusing on linguistic features alone, since it would take longer to detect propaganda than to generate it. Thus, in the future it will be necessary to go beyond texts and to also analyze the network nodes and the connectivity patterns through which propaganda spreads. 
    \item Spreading  propaganda through multiple modalities is  increasingly popular. Maliciously crafted images or videos can be more effective than articles when targeting  the millennial generation, who is more familiar with watching than reading. Again, research in detecting propaganda needs to move beyond text analysis, and to embrace more comprehensive analyses that span over various data modalities.

\end{enumerate}

\section{Conclusion\label{sec:conclusion}}

Among the contributions of our work, we  surveyed state-of-the-art computational propaganda detection methodologies. We also showed how the rapid pace of evolution of the techniques adopted by an adversary are impairing current propaganda detection solutions. 
Further, we justified our call for moving beyond textual analysis and we argued for the need of combined efforts blending  Natural Language Processing, Network Analysis, and Machine Learning.
Finally, we showed concrete promising research directions in the field of computational propaganda detection.

\bibliographystyle{named}
\bibliography{ijcai20}

\end{document}